\documentclass[sigconf]{acmart}

\renewcommand\footnotetextcopyrightpermission[1]{}
\fancypagestyle{plain}{%
  \fancyhf{}  

}

\fancypagestyle{firstpagestyle}{%
  \fancyhf{}

}
\makeatletter
\def\@conferenceinfo{}
\def\@acmConference{}
\makeatother

\usepackage{calrsfs}
\usepackage{amsmath}
\usepackage{multirow}
\usepackage{array}
\usepackage{makecell}

\AtBeginDocument{%
  }

\setcopyright{acmlicensed}
\copyrightyear{2025}
\acmYear{2025}
\acmDOI{XXXXXXX.XXXXXXX}
\acmConference[Seg-Wild]{Make sure to enter the correct
  conference title from your rights confirmation email}{2025}{07}
\acmISBN{978-X-XX-XXXX-X/YY/MM}

\begin{document}
\title{Seg-Wild: Interactive Segmentation based on 3D Gaussian Splatting for Unconstrained Image Collections}


\author{Yongtang Bao}
\affiliation{%
 \institution{College of Computer Science and Engineering, Shandong University of Science and Technology}
 \city{Qingdao}
 \country{China}}
\email{baozi0221@sdust.edu.cn}

\author{Chengjie Tang}
\affiliation{%
 \institution{College of Computer Science and Engineering, Shandong University of Science and Technology}
 \city{Qingdao}
 \country{China}}
\email{tangcj725@sdust.edu.cn}

\author{Yuze Wang\textsuperscript{*}}
\affiliation{%
  \institution{State Key Laboratory of Virtual Reality Technology and Systems, Beihang University}
  \state{Beijing}
  \country{China}}
\email{wangyuze1998@buaa.edu.cn}

\author{Haojie Li\textsuperscript{*}}
\affiliation{%
 \institution{College of Computer Science and Engineering, Shandong University of Science and Technology}
 \city{Qingdao}
 \country{China}}
\email{hjli@sdust.edu.cn}

\thanks{\textsuperscript{*}Corresponding authors}

\renewcommand{\shortauthors}{Bao et al.}


\begin{abstract}
Reconstructing and segmenting scenes from unconstrained photo collections obtained from the Internet is a novel but challenging task. Unconstrained photo collections are easier to get than well-captured photo collections. These unconstrained images suffer from inconsistent lighting and transient occlusions, which makes segmentation challenging. Previous segmentation methods cannot address transient occlusions or accurately restore the scene's lighting conditions. Therefore, we propose Seg-Wild, an interactive segmentation method based on 3D Gaussian Splatting for unconstrained image collections, suitable for in-the-wild scenes. We integrate multi-dimensional feature embeddings for each 3D Gaussian and calculate the feature similarity between the feature embeddings and the segmentation target to achieve interactive segmentation in the 3D scene. Additionally, we introduce the Spiky 3D Gaussian Cutter (SGC) to smooth abnormal 3D Gaussians. We project the 3D Gaussians onto a 2D plane and calculate the ratio of 3D Gaussians that need to be cut using the SAM mask. We also designed a benchmark to evaluate segmentation quality in in-the-wild scenes. Experimental results demonstrate that compared to previous methods, Seg-Wild achieves better segmentation results and reconstruction quality. Our code will be available at https://github.com/Sugar0725/Seg-Wild.
\end{abstract}



\maketitle

\section{Introduction}
With the development of 3D scene representation technology, scene segmentation from reconstructed scenes has become increasingly popular in computer vision and computer graphics. Recently, researchers have conducted extensive studies on scene reconstruction and 3D scene perception. Neural Radiance Field (NeRF) ~\cite{nerf, tensorf, scarf, mipnerf} has significantly advanced 3D scene reconstruction by implicitly representing scenes. However, the high computational cost of NeRF remains a major obstacle to its widespread adoption. 3D Gaussian Splatting (3DGS)~\cite{3dgs} has gradually emerged as a mainstream static 3D scene modeling approach. 3DGS represents the point cloud data in a scene by distributing 3D Gaussians in 3D space. It uses optimizable Gaussian parameters, such as position, size, rotation, color, and opacity, to describe continuous 3D structures, thus enabling more flexible view synthesis and highly efficient GPU-accelerated computations. Based on 3DGS, researchers have carried out extensive investigations~\cite{look, feature3d}, which has further broadened the application scope of 3DGS.

\begin{figure}[H]
\centering
\includegraphics[width=0.7\columnwidth]{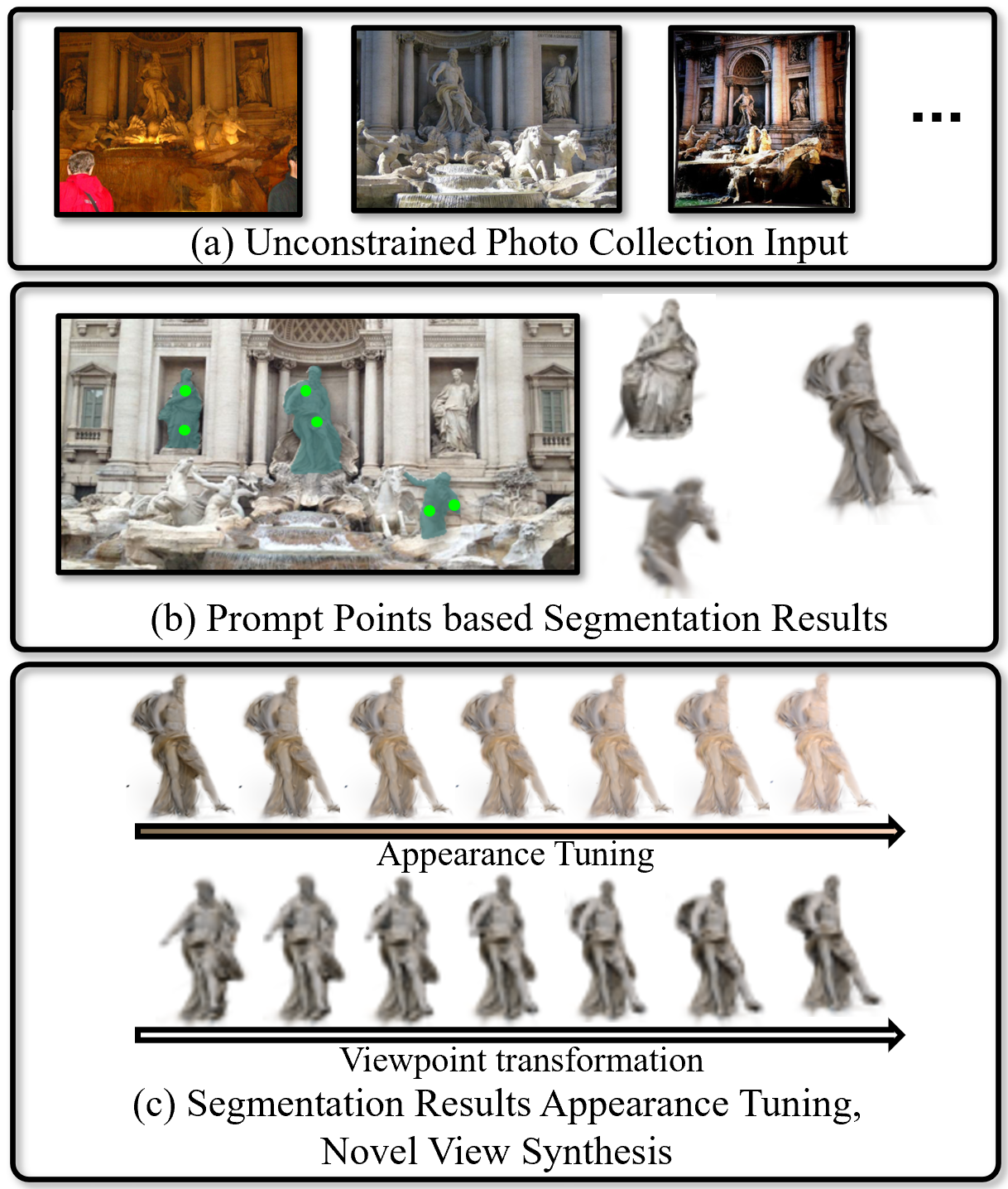}
\caption{(a) Our method uses unconstrained photo collections as input. (b) Our segmentation framework is optimized to address the issues of transient occlusions and inconsistent lighting conditions in in-the-wild scenes, producing high-quality interactive segmentation results. (c) Our segmentation also supports novel view synthesis and appearance tuning.}
\label{fig:teaser}
\end{figure}

3D scene segmentation based on volume rendering has recently attracted growing attention. Obtaining selectable and fine-grained 3D segmentation from reconstructed 3D scenes remains challenging. SAGA~\cite{saga} and SA3D~\cite{sa3d} utilize 3D Gaussian Splatting (3DGS)~\cite{3dgs} and NeRF~\cite{nerf} as 3D priors, respectively. These methods embed semantic features into 3D Gaussians or neural radiance fields and optimized them with other parameters to improve 3D segmentation efficiency and granularity. B Dou \textit{et al.}~\cite{doubin} backproject DINO multi-scale semantic features from 2D images onto the 3D Gaussian point cloud. They also extract geometric features using a spatial point cloud network and design a dual-feature fusion module that transitions from global to local representations to generate compact segmentation labels. 
WildSeg3D~\cite{wildseg} employs a feed-forward design for real-time 2D-to-3D interactive segmentation without scene-specific training, through three stages: (1) generating multi-view masks with SAM2~\cite{sam2}; (2) aligning point clouds via Dynamic Global Aligning (DGA); and (3) mapping masks to 3D using Multi-view Group Mapping (MGM). However, all these methods rely on well-captured photo collections with consistent lighting and no transient occlusions. It results in high costs for collecting such photo collections and limits the practical application of these methods. Recently, some approaches~\cite{look, wildgaussians, wegs, gsw} have initiated the exploration of reconstructing 3D outdoor scenes using unconstrained photo collections based on 3DGS. During training, they concurrently optimize consistent appearance embeddings and transient occlusion masks for each image. This concurrent optimization addresses view sparsity and transient occlusions in in-the-wild scene reconstruction. However, given the formidable challenges in semantic feature embeddings and refined segmentation within in-the-wild scenes, there are still scarce studies on scene segmentation methods grounded in unconstrained photo collections.

Therefore, we propose an interactive 3D segmentation method based on 3D Gaussian Splatting for unconstrained image collections (Seg-Wild). It adapts to scenes reconstructed from such photo collections and enables efficient, accurate segmentation. Seg-Wild extracts image feature embeddings using the 2D segmentation foundation model SAM~\cite{sam} and integrates a feature embedding module into the 3DGS training pipeline to lift 2D features into 3D space. Specifically, we optimize the 3D affinity feature by minimizing a 2D feature loss between the rendered image and the ground truth. Additionally, we constrain feature aggregation using SAM’s global segmentation results. The segmentation task then selects relevant 3D Gaussians by comparing the target object’s features with those embedded in each 3D Gaussian. The initial selection may include spiky 3D Gaussians that extend beyond object boundaries. To address this, we introduce a 3D Gaussian cutter that splits these 3D Gaussians and removes sharp regions, improving accuracy while preserving essential components. Moreover, we observed that a fixed scale in SAM’s global segmentation leads to inconsistent granularity, so we further optimize the segmentation prompts by designing a SAM segmentation scale adaptation module that accounts for the depth of 3D Gaussians.

The main contributions of this paper are summarized as follows:
\begin{itemize}
    \item We propose a novel 3D segmentation pipeline that is adaptable to unconstrained datasets. It enables both 3D reconstruction and segmentation from unconstrained photo collections, thereby reducing data acquisition costs.
    \item A scale-adaptive segmentation module (SASM) for SAM, which considers the depth and density of 3D Gaussians, efficiently optimizing SAM’s global segmentation results and providing more accurate constraints for the feature embedding process.
    \item A spiky 3D Gaussian cutter (SGC) that smooths segmentation results, removes spiky regions, retains 3D Gaussians contributing to the segmentation, and improves 3D segmentation accuracy.
    \item We design a benchmark for quantitative analysis. Experiments demonstrate that Seg-Wild underwent comprehensive evaluation and showed strong segmentation performance.
\end{itemize}

\section{Related Work}
\subsection{Segmentation in 3D Scenes}
With the development of 3D scene reconstruction methods~\cite{ripnerf, sg_nerf, mip3dgs, evolving}, research on segmentation within 3D scenes has gained significant attention. Some researchers have utilized the spatial information from point clouds in these scenes for segmentation and achieved remarkable results~\cite{pointnet, pointnet++, largepoint}. With the advent of Neural Radiance Field (NeRF)~\cite{nerf} and 3D Gaussian Splatting (3DGS)~\cite{3dgs}, which have improved the quality of visualization and reconstruction, more researchers~\cite{goi, drag, wildseg} have turned their focus to segmentation within scenes reconstructed by NeRF and 3DGS. ISRF~\cite{isrf} deliverd high-precision, real-time segmentation of complex 3D objects in a voxelized radiance field via interactive stroke guidance, bidirectional search in a spatial-semantic space, and DINO~\cite{dino,dinov2} semantic distillation. SA3D~\cite{sa3d} performd 3D segmentation of open scenes using reverse mask rendering of NeRF and a cross-view self-prompting mechanism, requiring only a single viewpoint and no 3D annotations.

Methods based on NeRF require a large amount of computational resources and time. Therefore, researchers have started to use 3DGS for segmentation tasks. CoSSegGaussians~\cite{doubin} fused DINO’s semantic features with the spatial geometry of 3D Gaussians using a feature fusion network and multi-scale aggregation, achieving efficient 3D zero-shot segmentation in open scenes. SAGA~\cite{saga} combined SAM’s 2D segmentation with 3D Gaussian point clouds by embedding multi-level features via contrastive learning, enabling real-time interactive 3D segmentation with multimodal prompts. Click-Gaussian~\cite{clickgs} constructed a 3D Gaussian semantic representation consistent across views using dual-level feature fields and Global Feature Guidance Learning (GFL). Gaussian Grouping~\cite{gaussiangrouping} integrated learnable identity encoding with SAM-based 2D segmentation features to achieve fully automatic, unsupervised instance segmentation in open-world 3D Gaussian scenes.

\subsection{3D Scene Representation from Unconstrained Photo Collections}
Reconstructing 3D scenes from unconstrained photo collections, such as internet-sourced images of famous landmarks, significantly reduces the cost of data acquisition. Researchers extensively explored scene representation using these diverse image sets~\cite{rome, neuralrendering, pt, factorized}. However, unconstrained photo collections pose various challenges, including inconsistent lighting conditions, differing color calibrations, and potential transient occlusions. These factors introduce significant difficulties in achieving accurate and reliable 3D scene reconstruction. Researchers made progress in traditional graphics-based reconstruction methods~\cite{online, neural3d, vis2mesh, pointdreamer, points2surf}. In volume rendering, NeRF-W \cite{nerfw} was the first to merge NeRF with unconstrained photo collections for 3D scene building. Using latent appearance embeddings and transient object decomposition, it addressed lighting and occlusion issues present in such collections for NeRF. K-planes~\cite{kplanes} decomposed the radiance field into low-rank planar tensor factors of space-time appearance. This representation enabled efficient modeling of dynamic scenes and appearance changes, achieving superior modeling of spatial and temporal changes while maintaining fast rendering performance.

With the development of 3DGS, which has higher rendering and reconstruction efficiency, researchers began to consider using 3DGS for reconstructing 3D scenes from unconstrained photo collections. GS-W \cite{gsw} proposed an adaptive 3D Gaussian method. It captures dynamic appearances from multi-layer features for novel view synthesis from unconstrained photos. WE-GS \cite{wegs} used a spherical harmonic coefficient module and a spatial attention mechanism to model lighting, optimize masks and appearance, balancing efficiency and quality. Wild-GS \cite{wildgs} enhanced geometric optimization and occlusion handling through hierarchical appearance modeling that fused material, illumination, and reflection properties. It further improved detail alignment by incorporating depth regularization and mask refinement. In addition, WildGaussians~\cite{wildgaussians} and SLS~\cite{sls} leveraged the vision foundation model DINO to extract feature maps and mitigate the impact of transient occlusions by decoupling dynamic regions based on DINO-derived features.

The researchers mentioned above have achieved remarkable progress in 3D scene reconstruction. However, we found that after in-the-wild scene reconstruction, there is no targeted segmentation work as an application of scene reconstruction. Therefore, our work proposes further developing segmentation work after scene representation from unconstrained photo collections, expanding the downstream tasks after scene reconstruction. Our method reduces the need for well-captured photo collections, lowers data acquisition costs, and achieves satisfactory segmentation results.

\section{Preliminaries}
3D Gaussian Splatting (3DGS)~\cite{3dgs} represents a static scene using a set of 3D anisotropic Gaussians $\mathcal{G} = \{ G_i(p_i, \Sigma_i, c_i, \alpha_i) \}_{i=1}^N$ and achieves real-time image rendering through a differentiable tile-based rasterizer. Each 3D Gaussian $G_i$ stores the positional information $p_i\in\mathbb{R}^3$ of the Gaussian center, 3D covariance information $\Sigma_i$ (which contains a rotation matrix $R_i\in\mathbb{R}^{3\times3}$ expressed in quaternions and a transformation matrix $t_i\in\mathbb{R}^3$) and opacity $\alpha_i\in\mathbb{R}$. We provide further details on implementation in the supplementary material.

The color information $c$ contains the base color information $Y_0\in\mathbb{R}^3$ that represents the zero-order spherical harmonic function and the spherical harmonic coefficients $Y_3\in\mathbb{R}^{45}$ that can represent up to third order. The essential technique of 3DGS is to project $N_\mathcal{G}$ 3D Gaussians onto a 2D image, parallel rasterization with CUDA for fast per-pixel $uv$ color rendering:
\begin{equation}
C\left(uv\right)=\sum_{i=1}^{N_\mathcal{G}}{c_i\alpha_i\left(p\right)}\prod_{j=1}^{i-1}{(1-\alpha_j\left(p\right))},
\label{eq5}
\end{equation}
where $\alpha_i\left(p\right)$ represents the opacity of the 3D Gaussian $G_i$ at the three-dimensional coordinate $p$.

The reconstruction part of our segmentation method is based on GS-W~\cite{gsw}, which adds dynamic appearance features modeling based on 3DGS. It uses a UNet to extract projected feature maps from reference images, capturing the global illumination and texture information at corresponding positions in the reference images. In addition, the UNet also generates a 2D binary mask $VM$ to distinguish the static scene from transient occluders and optimize the training loss. The loss function formula for transient objects is as follows:
\begin{equation}
L_{vm}=L_2\left(VM,\ 1\right).
\label{eq6}
\end{equation}

Similar to 3DGS~\cite{3dgs}, the color loss function $L_c$ of GS-W applies three types of loss functions: $L_1$ , $L_{SSIM}$~\cite{ssim}, and $L_{LPIPS}$~\cite{lpips}. The overall color loss function calculates the difference between the ground truth image $I_{gt}$ and the rendered image $I_r$ to optimize the 3D Gaussian scene. The formula is as follows:
\begin{equation}
\begin{split}
L_c &= \lambda_1 L_1(VM \odot I_r, VM \odot I_{gt}) \\
    &+ \lambda_2 L_{SSIM}(VM \odot I_r, VM \odot I_{gt}) \\
    &+ \lambda_3 L_{LPIPS}(I_r, I_{gt}),
\end{split}
\label{eq7}
\end{equation}
where $\odot$ represents the Hadamard product, and $\lambda_1$, $\lambda_2$ and $\lambda_3$ represent the loss weights of $L_1$, $L_{SSIM}$ and $L_{LPIPS}$, respectively.

\begin{figure*}[htb]
  \centering
  \includegraphics[width=0.8\linewidth]{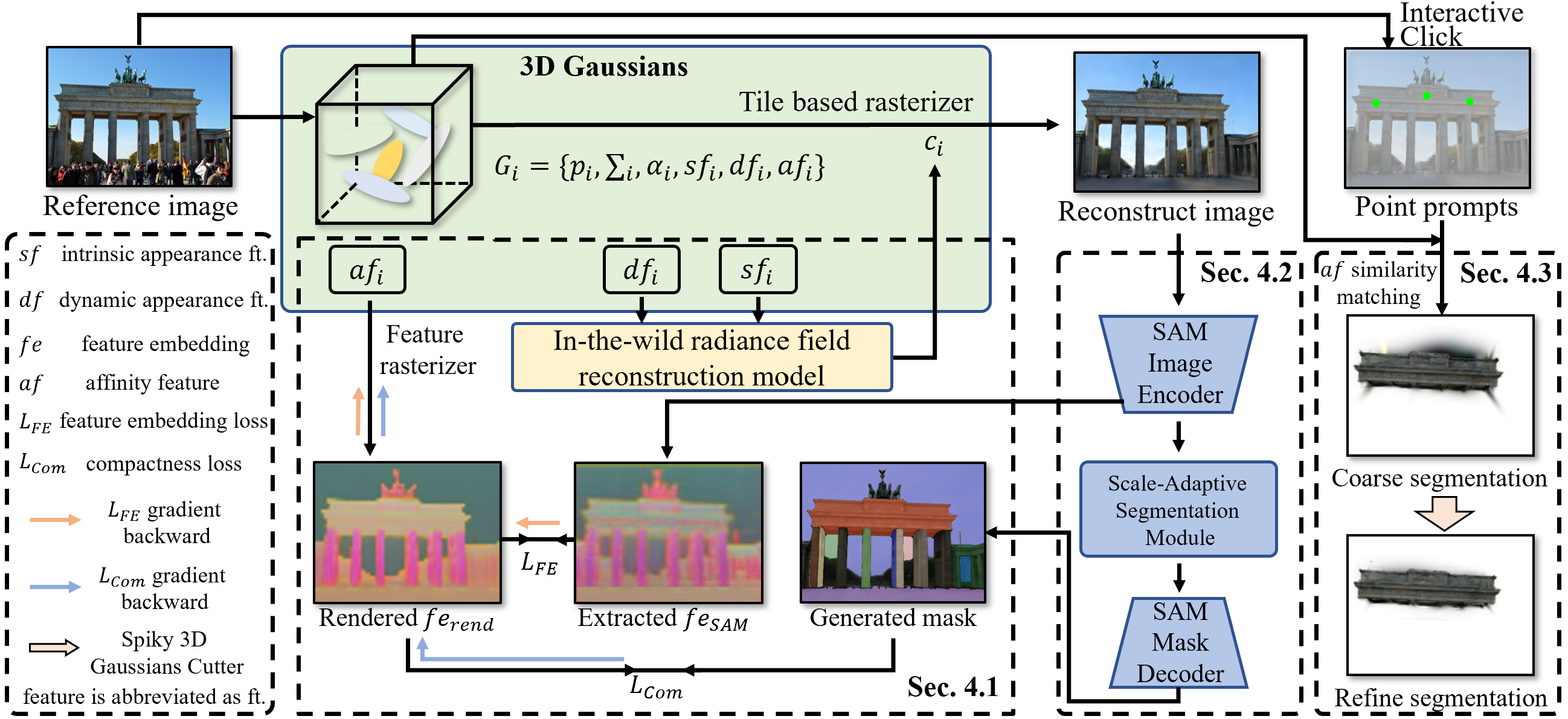}
  \parbox[t]{1\columnwidth}{\relax}
  \caption{An overview of our framework. During the reconstruction of in-the-wild scenes, we embed affinity features into 3D Gaussians to construct a 3D feature field. Optimized by the scale-adaptive segmentation module (SASM), the SAM mask promotes feature compactness for improved segmentation. In the segmentation process, we use the prompt points $\mathscr{pp}$ to find the feature embeddings of the reference image $I_i$ and calculate the similarity with the affinity features $af$ to identify the similar 3D Gaussians. The spiky 3D Gaussians cutter (SGC) refines the segmentation results to obtain the final output segmentation.}
    \label{fig:pipline}
\end{figure*}

\section{Method}
Given a set of unconstrained photo collections $I=\{I_1,I_2,\ldots,I_N\}$ from the Internet, which suffer from inconsistent lighting conditions, varying post-processing effects, and transient occlusions, our method aims to obtain accurate feature embeddings by optimizing the segmentation scale of the SAM model, thereby achieving precise interactive 3D segmentation. 

Figure~\ref{fig:pipline} shows the overall workflow of our method. First, we extract feature embeddings using the Segment Anything Model (SAM) and embed them into each Gaussian based on GS-W. We elevate the 2D features to 3D affinity features as a reference for segmentation. The masks generated by SAM enhance the feature consistency (Section~\ref{2Dto3D}). Second, to address the inconsistency of SAM-generated masks in unconstrained photo collections, we design a scale-adaptive segmentation module (SASM) that considers the depth and density of 3D Gaussians to optimize SAM segmentation (Section~\ref{SASM}). Finally, we introduce a spiky 3D Gaussians cutter (SGC) that can smooth the segmentation results and remove the spiky parts (Section~\ref{SGC}). For additional details, please refer to the supplementary material.

\subsection{2D Features to 3D Features}
\label{2Dto3D}
Our interactive segmentation method identifies 3D Gaussians in the scene by matching their affinity features with the target objects. This process relies on similarity comparison, making constructing a 3D feature field crucial for accurate segmentation. We adjusted the data structure of 3D Gaussians to $G_i = \{p_i, \Sigma_i, \alpha_i, sf_i, df_i, af_i\}$ ($p_i$ represents the center coordinate of $G_i$, and $\Sigma_i$ represents the three-dimensional covariance matrix of $G_i$. $\alpha$ represents the opacity of $G_i$. ${sf}_i$ and ${df}_i$ are the intrinsic and dynamic appearance features of $G_i$, respectively. Section A.2 of the supplementary material provides detailed definitions and motivation for these two features. $af_i$ represents the affinity feature of $G_i$, used to construct the 3D feature field). We train the 3D feature field simultaneously while reconstructing the 3D scene. We realize the 3D feature field by optimizing the loss between the 2D feature embeddings generated by SAM and the render feature embeddings obtained from feature rendering in the 3D scene.

Next, we will elaborate on the construction process of the 3D feature field. We obtain the 2D feature embeddings ${fe}_{SAM}\in\mathbb{R}^{H^\prime\times W^\prime\times C}$ for each image through the pre-trained image encoder of SAM. Based on empirical observations and computational considerations, we set the feature dimension $C$ to 64 by compressing the 256-dimensional ${fe}_{SAM}$ using PCA. Compression details are provided in the supplementary material. The rendering of the feature embeddings ${fe}_{rend}$ for a specified view follows a similar process to RGB image rendering in 3DGS. The affinity feature ${af}_i$ of each 3D Gaussian is projected onto the 2D plane to obtain ${fe}_{rend}$ using CUDA-accelerated rasterization. The following equation defines the process:
\begin{equation}
{fe}_{rend}=\sum_{u=0}^{W-1}\sum_{v=0}^{H-1}{\sum_{i=1}^{N_\mathcal{G}}{{af}_i\alpha_i\left(u,v\right)}\prod_{j=1}^{i-1}{(1-\alpha_j\left(u,v\right))}},\ 
\label{eq9}
\end{equation}
where $u$ and $v$ represent the horizontal and vertical coordinates in the 2D image, respectively, and $N_\mathcal{G}$ denotes the number of 3D Gaussians.

After obtaining the feature embeddings ${fe}_{SAM}$ extracted by the 2D foundation model SAM and the rendered feature embeddings ${fe}_{rend}$, we calculate the $L_1$ loss of each pixel as the feature embeddings loss $L_{FE}$:
\begin{equation}
L_{FE} = \| f e_{SAM} - f e_{rend} \|_1.
\label{eq10}
\end{equation}

Our experiments found that training the 3D feature field using only the feature embeddings loss $L_{FE}$ leads to loosely clustered affinity features within the 3D scene, meaning the feature boundaries between objects are unclear. To enhance the compactness of the affinity features, we draw inspiration from the correspondence loss proposed in SAGA~\cite{saga}. Specifically, we use the SAM global segmentation mask as a reference to explicitly reinforce the feature boundaries between different objects.

First, we use SAM to generate a series of masks for each image as segmentation priors to measure the regional similarity between pixels. Section ~\ref{SASM} provides a detailed explanation of the segmentation process. We calculate the Intersection over Union (IoU) similarity matrix $S({px}_i, {px}_j)$ between the masks $M_{{px}_i}$ and $M_{{px}_j}$ corresponding to two pixels ${px}_i$ and ${px}_j$. 

Second, we calculate the cosine similarity between the two pixels $px_i$ and $px_j$ in the rendered feature embeddings ${fe}_{rend}$, forcing the network to reduce the feature differences in the regions with similar feature embeddings. The specific formula is as follows:
\begin{equation}
C({px}_i,{px}_j)=max(0,<{fe}_{rend}^{{px}_i},{fe}_{rend}^{{px}_j}>),
\label{eq12}
\end{equation}
where ${fe}_{rend}^{{px}_i}$ represents the feature embeddings corresponding to pixel ${px}_i$, and $<\cdot,\cdot>$ denotes the cosine similarity computation.

Finally, the compactness loss $L_{Com}$ is defined as:
\begin{equation}
L_{Com}=\sum_{{px}_i}^{H\times W}\sum_{{px}_j}^{H\times W}{S({px}_i,{px}_j)C({px}_i,{px}_j)},
\label{eq13}
\end{equation}
where $H$ and $W$ represent the height and width of each image $I$, respectively.

We facilitate downstream segmentation tasks by introducing $L_{FE}$ and $L_{Com}$, which embed compact affinity features $af$ into in-the-wild scenes. Our loss function jointly optimizes both the 3D scene representation and the 3D feature field. The detailed formulation is defined as follows:
\begin{equation}
L=L_C+\lambda_{FE}L_{FE}+\lambda_{Com}L_{Com}, 
\label{eq14}
\end{equation}
where the calculation of $L_C$ is as shown in Eq.~\ref{eq7}, and $\lambda_{FE}$ and $\lambda_{Com}$ are 0.7 and 0.3, respectively.

\textbf{\begin{figure}[H]
\centering
\includegraphics[width=0.9\columnwidth]{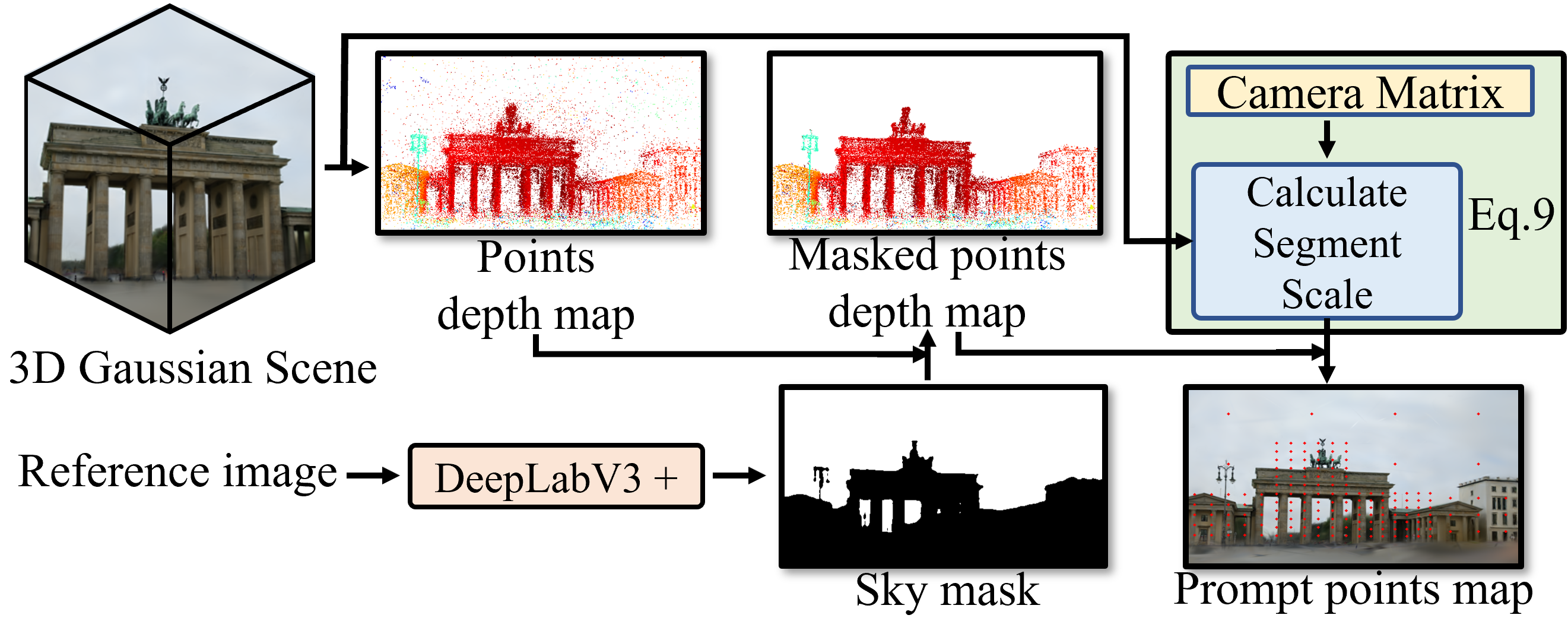}
\caption{The procedure for SASM is to generate prompt points to optimize the segmentation scale. We use DeepLabV3+ to extract the sky mask. Given a point depth map projected from the 3D scene onto a 2D plane, we apply the sky mask to remove the sky region and prevent sky depth from affecting the sampling distribution. We divide the reference image into regular rectangular blocks, calculate the total depth in each block, and allocate sampling points based on these values.}
\label{fig:SASM}
\end{figure}}

\subsection{Scale-Adaptive Segmentation Module (SASM)}
\label{SASM}
We detail the reconstruction process of the 3D feature field in Section~\ref{2Dto3D}. During the computation of $L_{Com}$, SAM segmentation masks are used as priors. However, due to the wide variation in camera distances within unconstrained photo collections, we observed that using a fixed segmentation scale for SAM leads to inconsistent granularity across different images. To address this, we design a scale-adaptive segmentation module (SASM), as shown in Figure~\ref{fig:SASM}, which considers camera distance and the spatial distribution of 3D Gaussians.

First, to enable SAM to adaptively adjust the segmentation scale according to the distance between the camera and the scene, we calculate the average distance from the camera to the center coordinates of 3D Gaussians in the 3D scene to obtain a relative scale suitable for SAM segmentation. Specifically, given a reference image $I$, we calculate the average distance $d_I$ from the projection of the center coordinates of all 3D Gaussians on the camera optical axis $o_I$ to the camera center point ${cc}_I$ as the distance parameter. The formula is defined as follows:
\begin{equation}
d_I=\frac{1}{N_\mathcal{G}}\sum_{j=1}^{N_\mathcal{G}}{(p_j-{cc}_I)\cdot o_I}, 
\label{eq15}
\end{equation}
where $N_\mathcal{G}$ represents the number of 3D Gaussians, and $p_j$ represents the center coordinate of the $j$-th 3D Gaussian. We normalize the distance value $d_I$ to the range [0, 1] and apply a linear transformation to map it to the range [4, 8], after which the result is rounded down to obtain the final segmentation scale ${SegS}_I$.

Second, we found that segmenting the image using uniform sampling points in the image would cause SAM to spend more attention on segmenting the background. Therefore, we consider using the depth map of 3D Gaussians as the basis for the distribution of adaptive sampling points. Areas that are close to the camera and have a sparse distribution of 3D Gaussians should be assigned fewer sampling points, while areas that are far from the camera and have a dense distribution of 3D Gaussians should be assigned more sampling points. We first project the center coordinate $p_i={(x_i,y_i,z_i)}^T$ of each 3D Gaussian onto the plane of the image $I$ to obtain $(u_i,v_i)$:
\begin{equation}
K [R \mid t] \begin{bmatrix} x_i \\ y_i \\ z_i \end{bmatrix} = K \begin{bmatrix} x_i^c \\ y_i^c \\ z_i^c \end{bmatrix} = \begin{bmatrix} u'_i \\ v'_i \\ w'_i \end{bmatrix},
\label{eq17}
\end{equation}
where ${(x_i^c, y_i^c, z_i^c)}^T$ is the 3D point in the camera coordinate system, $K$ is the intrinsic matrix, $R$ represents the rotation matrix, and $t$ represents the translation matrix. We then generate the points depth map ${DM}_I$ corresponding to the image $I$. The specific formula for calculating the depth value ${DM}_I(u_i,v_i)$ of each pixel is as follows:
\begin{equation}
\begin{aligned}
DM_I(u_i, v_i) &= \min(DM_I(u_i, v_i), Z(u_i, v_i)),
\\
(u_i, v_i) &= \left( \frac{{u'}_i}{{w'}_i}, \frac{{v'}_i}{{w'}_i} \right),
\end{aligned}
\label{eq19}
\end{equation}
where $Z(u_i,v_i)$ represents the previously minimum depth value at $(u_i,v_i)$. We select the minimum depth among all 3D Gaussian projections at $(u_i,v_i)$ and add it to ${DM}_I$. After generating ${DM}_I$, we observe that unconstrained image sets often contain sky regions with tremendous depth values. Such high values contradict our assumption that greater depth should correspond to denser sampling. To address this problem, we suppress the depth values of sky regions to focus more on the segmentation of foreground objects. We use the pre-trained DeepLabV3+~\cite{deeplabv3+, mobilenets, cityscapes} model as the sky mask extractor to identify the sky (LSeg~\cite{lseg} and SEEM~\cite{seem} can also be used as sky mask extractors). We combine the sky mask ${SM}_I$ and the depth map ${DM}_I$ to generate the depth map ${DM'}_I$ with the sky filtered out:
\begin{equation}
{DM'}_I=(1-{SM}_I)\odot{DM}_I+{SM}_I\odot{\min(DM}_I).
\label{eq20}
\end{equation}

Finally, we utilize the segmentation scale ${SegS}_I$ and the filtered depth map ${DM'}_I$, in which the sky regions have been removed, to compute the prompt points map ${PPM}_I$ as input to SAM. The process is as follows. We first divide the depth map ${DM'}_I$ into a grid of size ${SegS}_I\times{SegS}_I$. For each grid cell $(i,j)$, we calculate the average depth $D_{(i,j)}$ using the following formula:
\begin{equation}
D_{(i,j)}=\frac{1}{\left|R_{(i,j)}\right|}\sum_{(u,v)\in R_{(i,j)}}{{DM'}_I(u,v), }
\label{eq21}
\end{equation}
where $R_{(i,j)}$ denotes the $(i,j)$-th grid region. Based on the average depth $D_{(i,j)}$ of each grid, we then compute the number of prompt points assigned to that grid, denoted as ${NPP}_{(i,j)}$:
\begin{equation}
NPP_{(i,j)} = \min(20, \max(1, \left\lfloor D_{(i,j)}\right\rfloor)).
\label{eq22}
\end{equation}
We then uniformly generate ${NPP}_{(i,j)}\times{NPP}_{(i,j)}$ prompt points within the grid to generate the regional prompt points map ${PPM}_I^{(i,j)}$:
\begin{equation}
\begin{aligned}
{PPM}_I^{(i,j)} = \Big\{ 
  & u_0 + \frac{(2m+1)}{2\cdot{NPP}_{(i,j)}} w_c,\ 
    v_0 + \frac{(2n+1)}{2\cdot{NPP}_{(i,j)}} h_c\ \Big|\ \\
  & m,n \in \left[0, {NPP}_{(i,j)} - 1 \right]
\Big\},
\end{aligned}
\label{eq23}
\end{equation}
where $w_c=\frac{W}{{SegS}_I}$ represents the width of each grid, $W$ is the width of the image. Similarly, $h_c=\frac{h}{{SegS}_I}$ represents the height of each grid, with $H$ denoting the height of the image. 

At last, all grid-level prompt points are aggregated to form the complete prompt points map.

\subsection{Spiky 3D Gaussian Cutter (SGC)}
\label{SGC}
After constructing the 3D feature field, we introduce the segmentation process. We first click on the target object to generate $N_{\mathscr{pp}}$ prompt points $\mathscr{pp}=\{(u_i,v_i)|i=1,\dots,N_p\}$ during the interactive segmentation process. We then use Eq.~\ref{eq9} to render the feature embeddings ${fe}_{rend}$, and find the corresponding feature embeddings ${fe}_{rend}(\mathscr{pp})\in\mathbb{R}^{N_{\mathscr{pp}}\times C}$ of the prompt points $\mathscr{pp}$ on ${fe}_{rend}$. We use the cosine similarity $s$ between ${fe}_{rend}(\mathscr{pp})$ and the affinity feature $af$ of 3D Gaussians as the preliminary criterion for determining whether 3D Gaussians are included in the segmentation result. The calculation formula is as follows:
\begin{equation}
s_i = \frac{fe_{rend}(u_i, v_i) \cdot a_f^T}{\| fe_{rend}(u_i, v_i) \| \| a_f \|},
\label{eq25}
\end{equation}
where $(u_i,v_i)$ represents the coordinates of the $i$-th prompt point, and $s_i\in\mathbb{R}^{N_\mathcal{G}}$ represents the similarity between the $i$-th prompt point and all 3D Gaussians. To simplify the subsequent similarity comparison, we transform the similarities of $N_{\mathscr{pp}}$ prompt points into the fused similarity $s_{fus}$. The formula is defined as follows:
\begin{equation}
s_{fus}=\sum_{i=1}^{N_{\mathscr{pp}}}\frac{e^{s_i}}{\sum_{j=1}^{N_p}e^{s_j}}s_i.
\label{eq26}
\end{equation}
We finally use the prompt points to generate a SAM mask $M_{2D}\in{\{0,1\}}^{H\times W}$ and map it to the 3D scene to assist in segmentation, obtaining the segmentation result $M_{3D}$:
\begin{equation}
(u_j,v_j)=\pi_{cam}(p_j) ,
\label{eq27}
\end{equation}
\begin{equation}
M_{3D} = \{ p_j | s_{fus} > \tau, M_{2D}(u_j, v_j) = 1 \},
\label{eq28}
\end{equation}
where $(u_j,v_j)$ represents the coordinates of the $j$-th 3D Gaussian projected onto the 2D plane, $\pi_{cam}(\cdot)$ represents the conversion formula from 3D to 2D coordinates (you can refer to Formulas Eq.~\ref{eq17} and Eq.~\ref{eq19}), and $\tau$ is a threshold we set with a value of 0.5.

During the experiment, we found that although the segmentation result $M_{3D}$ shows good performance, the long axes of the ellipsoids of some slender 3D Gaussians whose centers are located within $M_{2D}$ protrude beyond the segmentation, which leads to sharp protrusions at the edges of the segmentation result. To smooth the segmentation result, we consider removing these spiky 3D Gaussians. However, directly removing these spiky 3D Gaussians may cause the segmentation result to miss key components in some cases. Therefore, we designed a spiky 3D Gaussian cutter (SGC) that can retain the part of the spiky Gaussians within $M_{2D}$ while cutting off the noise exposed outside the segmentation result. The specific process is as follows.  

First, given a spiky 3D Gaussian ${SG}_i$, we calculate the long axis of ${SG}_i$ in the 2D projection. We project the covariance matrix $\Sigma^i\in\mathbb{R}^{3\times3}$ onto the 2D plane to obtain the 2D covariance matrix $\Sigma_{uv}^i\in\mathbb{R}^{2\times2}$.
We calculate the maximum eigenvalue $\lambda_{max}$ and the corresponding eigenvector $v_{max}$ based on $\Sigma_{uv}^i$. According to Eq.~\ref{eq17} and Eq.~\ref{eq19}, we calculate the coordinates $(u_i,v_i)$ of the projection center point of ${SG}_i$ on the 2D plane. The calculation formulas for the 2D coordinates of the two endpoints ${uv}_1^i$ and ${uv}_2^i$ of the long axis are as follows:
\begin{equation}
{uv}_1^i=uv_i+\frac{3\sqrt{\lambda_{max}}}{2}\cdot v_{max}\ ,\ {uv}_2^i=uv_i-\frac{3\sqrt{\lambda_{max}}}{2}\cdot v_{max}.
\label{eq30}
\end{equation}

Second, we determine the proportion of the spiky 3D Gaussian that needs to be cut off by calculating the range of the long axis of ${SG}_i$ covered by the SAM mask $M_{2D}$. Given $N_{\mathscr{px}}$ intermediate pixels ${\mathscr{px}}_{mid}$ located on the straight line between ${uv}_1^i$ and ${uv}_2^i$, we calculate the coverage ratio $r_i$ of the pixels in the SAM mask $M_{2D}$:
\begin{equation}
r_i=\frac{\sum_{j=1}^{N_{\mathscr{px}}}{M_{2D}\left(\mathscr{px}_{mid}^j\right)}}{N_{\mathscr{px}}}.
\label{eq31}
\end{equation}

Finally, we use the coverage ratio $r_i$ to adjust 2D center coordinate $uv_i$ of the spiky 3D Gaussian ${SG}_i$ and the scaling scale ${scale}_i$ stored in the covariance matrix $\Sigma^i$ of ${SG}_i$:
\begin{equation}
\begin{split}
uv'_i &= uv_i + (1 - r_i)\sqrt{\lambda_{max}}v_{max}, scale'_i= r_i\,scale_i,
\end{split}
\label{eq32}
\end{equation}
where ${uv'}_i$ is the adjusted 2D center coordinate, and ${scale'}_i$ is the adjusted scaling scale. We back-project ${uv'}_i$ to 3D by inverting Eq.~\ref{eq17} to obtain the 3D center coordinate.

\begin{table*}[ht]
\centering
\setlength{\tabcolsep}{2.5pt}
\renewcommand{\arraystretch}{0.8}
\caption{We perform a quantitative comparison of segmentation performance (IoU and Acc) on three representative scenes from the Photo Tourism (PT) dataset~\cite{pt}. Each scene includes three segmentation targets. Compared with three methods, Feature 3DGS~\cite{feature3d} and SAGA~\cite{saga}, both limited by transient occlusions, and the GS-W~\cite{gsw} reconstruction method that employs projection-based segmentation, our method achieves superior results across all metrics and targets. The abbreviations ``BG'', ``TM'', and ``TF'' refer to the ``Brandenburg Gate'', the ``Taj Mahal'', and the ``Trevi Fountain'', respectively.}
\begin{tabular}{l l c c c c c c c c}
\toprule
\multirow{2}{*}{\centering\arraybackslash Scene} & 
\multirow{2}{*}{\centering\arraybackslash \parbox{2cm}{\centering \vspace{3pt}Segmentation\\ name}}&
\multicolumn{2}{c}{Feature 3DGS} & 
\multicolumn{2}{c}{SAGA} & 
\multicolumn{2}{c}{GS-W+Projection} &
\multicolumn{2}{c}{Seg-Wild (Ours)} \\
\cmidrule(lr){3-4} \cmidrule(lr){5-6} \cmidrule(lr){7-8} \cmidrule(lr){9-10}
& & IoU$\uparrow$ & Acc$\uparrow$ & IoU$\uparrow$ & Acc$\uparrow$ & IoU$\uparrow$ & Acc$\uparrow$ & IoU$\uparrow$ & Acc$\uparrow$\\
\midrule
\multirow{3}{*}{\begin{tabular}[c]{@{}l@{}}BG\end{tabular}} & Lintel & 0.911$\pm$0.002 & 0.926$\pm$0.004 & 0.913$\pm$0.001 & 0.933$\pm$0.005 & 0.803$\pm$0.001 & 0.852$\pm$0.003 & \textbf{0.941$\pm$0.003} & \textbf{0.974$\pm$0.003} \\
& Right pavilion & 0.787$\pm$0.003 & 0.865$\pm$0.005 & 0.840$\pm$0.003 & 0.913$\pm$0.006 & 0.709$\pm$0.004 & 0.814$\pm$0.005 & \textbf{0.958$\pm$0.004} & \textbf{0.971$\pm$0.002} \\
& Left pavilion & 0.770$\pm$0.001 & 0.843$\pm$0.002 & 0.851$\pm$0.004 & 0.911$\pm$0.002 & 0.687$\pm$0.002 & 0.800$\pm$0.004 & \textbf{0.892$\pm$0.001} & \textbf{0.961$\pm$0.005} \\
\midrule
\multirow{3}{*}{\begin{tabular}[c]{@{}l@{}}TF\end{tabular}} & Oceanus & 0.705$\pm$0.002 & 0.883$\pm$0.003 & 0.812$\pm$0.004 & 0.873$\pm$0.002 & 0.652$\pm$0.001 & 0.792$\pm$0.005 & \textbf{0.839$\pm$0.002} & \textbf{0.906$\pm$0.001} \\
& Right triton & 0.719$\pm$0.004 & 0.811$\pm$0.005 & 0.760$\pm$0.001 & 0.894$\pm$0.002 & 0.674$\pm$0.003 & 0.802$\pm$0.003 & \textbf{0.813$\pm$0.004} & \textbf{0.929$\pm$0.003} \\
& Left triton & 0.806$\pm$0.001 & 0.881$\pm$0.005 & 0.753$\pm$0.007 & 0.836$\pm$0.004 & 0.716$\pm$0.002 & 0.813$\pm$0.004 & \textbf{0.830$\pm$0.002} & \textbf{0.937$\pm$0.005} \\
\midrule
\multirow{3}{*}{\begin{tabular}[c]{@{}l@{}}TM\end{tabular}} & Onion dome & 0.834$\pm$0.003 & 0.882$\pm$0.004 & 0.853$\pm$0.002 & 0.896$\pm$0.001 & 0.801$\pm$0.001 & 0.849$\pm$0.002 & \textbf{0.924$\pm$0.001} & \textbf{0.969$\pm$0.004} \\
& Left minaret & 0.739$\pm$0.005 & 0.854$\pm$0.002 & 0.727$\pm$0.005 & 0.845$\pm$0.003 & 0.741$\pm$0.005 & 0.855$\pm$0.001 & \textbf{0.778$\pm$0.006} & \textbf{0.872$\pm$0.003} \\
& Right chhatris & 0.751$\pm$0.004 & 0.762$\pm$0.003 & 0.772$\pm$0.001 & 0.864$\pm$0.005 & 0.733$\pm$0.001 & 0.816$\pm$0.003 & \textbf{0.838$\pm$0.002} & \textbf{0.921$\pm$0.006} \\
\bottomrule
\end{tabular}
\label{tab:segmentation_comparison}
\end{table*}

\section{Experiments}
\subsection{Implementation Details}
We implemented our method using PyTorch~\cite{pytorch} and trained our network with the Adam optimizer~\cite{adam}. We trained the entire model for 70,000 iterations on an Nvidia RTX 4060 GPU. We performed feature extraction and SAM Mask generation during the training process, which took approximately 1.5 hours. During the training, we adopted other hyperparameter settings similar to those in 3DGS~\cite{3dgs}.

\subsection{Datasets}
We evaluate our method on two challenging datasets: the Photo Tourism (PT) dataset~\cite{pt} and the NeRF-On-the-go dataset~\cite{nerf_on_the_go}. The Photo Tourism (PT) dataset contains unconstrained Internet photo collections of famous landmarks like Trevi Fountain (TF), Taj Mahal (TM), and Brandenburg Gate (BG). Captured by different users under varying lighting, camera settings, and viewpoints, it introduces substantial variation and realism. Each scene includes 800–1500 images, making PT a strong benchmark for evaluating 3D reconstruction and segmentation in complex, unconstrained settings. The NeRF-On-the-go dataset supports mobile scene reconstruction and includes high-resolution videos captured via handheld smartphones while walking around targets. Around 200–300 frames are extracted per sequence, reflecting real-world motion, viewpoint changes, and lighting variations. It is a practical dataset for neural rendering and 3D segmentation in real-life scenes. We referenced several benchmark datasets~\cite{nvos, spinnerf} for mainstream segmentation tasks and selected representative scenes from the Photo Tourism (PT) dataset. Each scene includes multiple segmentation targets and is a benchmark for in-the-wild segmentation. We provide implementation details in the supplementary material.

\subsection{Quantitative Comparison}
In the quantitative evaluation, we compare our proposed method with SAGA~\cite{saga} and Feature 3DGS~\cite{feature3d}, using two widely adopted metrics in segmentation tasks: Intersection over Union (IoU) and Accuracy (Acc). As presented in Table~\ref{tab:segmentation_comparison}, Feature 3DGS and SAGA are tailored for well-captured photo collections and struggle to cope with the transient occlusions commonly found in unconstrained photo collections. As a result, the segmentation of both methods exhibits visual artifacts. Although SAGA outperforms Feature 3DGS due to its implementation of affinity feature compactness constraints, it remains inadequate in addressing the complex challenges inherent in in-the-wild scenes. In contrast, our method is purposefully designed to address the complexities of unconstrained photo collections. Introducing a series of targeted optimizations for real-world scenarios outperforms existing approaches, achieving the highest IoU and Acc scores across all benchmark scenes.

\subsection{Qualitative Comparison}
\begin{figure*}[ht]
\centering
\includegraphics[width=0.7\textwidth]{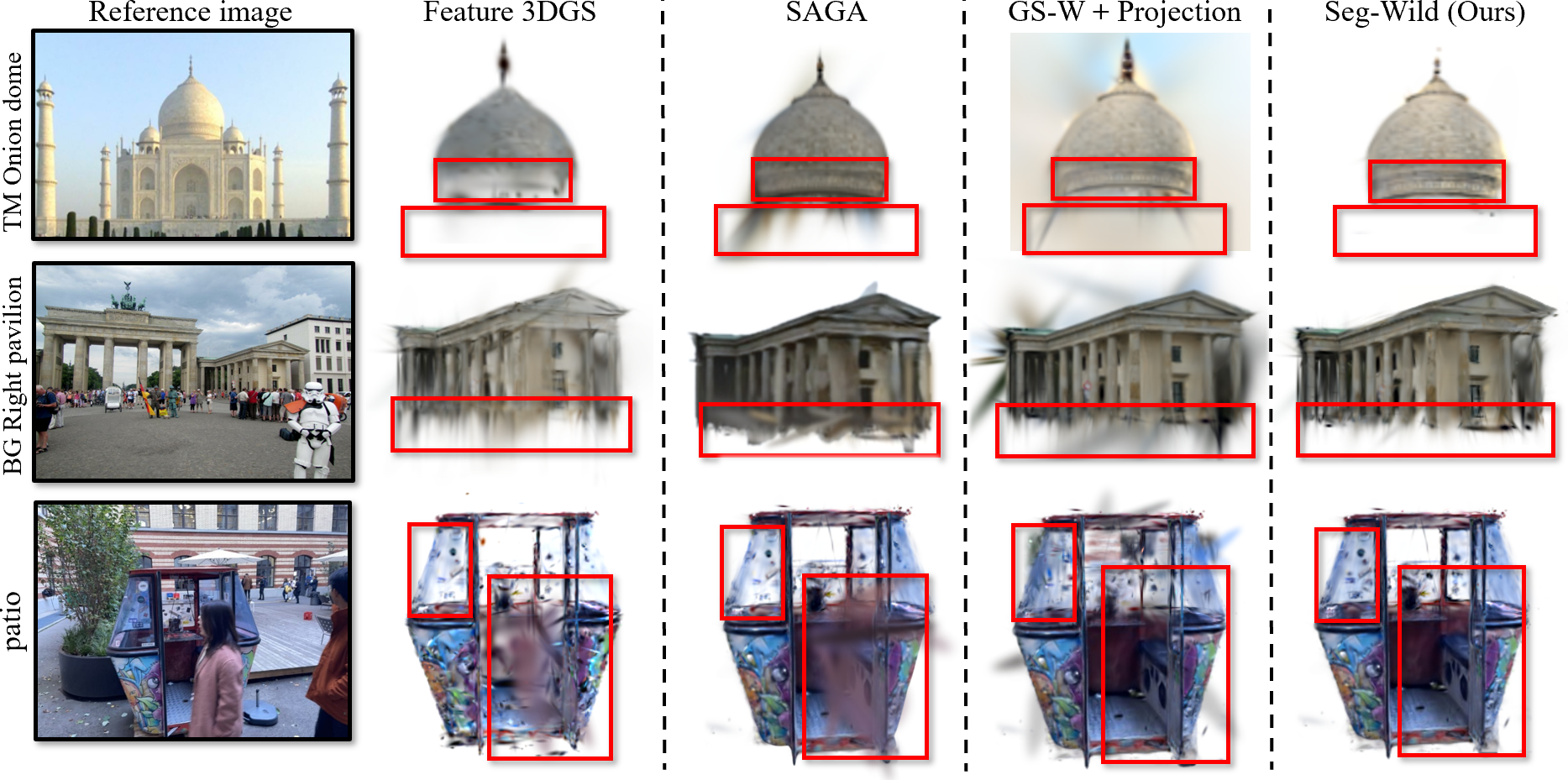}
\caption{We perform a qualitative comparison of segmentation results across four methods: Feature 3DGS~\cite{feature3d}, SAGA~\cite{saga}, GS-W~\cite{gsw} with projection-based segmentation, and our proposed approach. Feature 3DGS exhibits loosely clustered segmentations due to the lack of feature compactification. SAGA enforces affinity constraints but fails to handle transient occlusions, leading to floating and spiky 3D Gaussians. GS-W, with projection-based segmentation, struggles to remove background Gaussians behind targets. In contrast, our method produces cleaner and more semantically consistent segmentations in unconstrained photo collections.}
\label{fig:qualitative}
\end{figure*}

Our qualitative experiments compare three methods: Feature 3DGS\\~\cite{feature3d}, SAGA~\cite{saga}, and GS-W~\cite{gsw} with projection-based segmentation. The first column in Figure~\ref{fig:qualitative} shows the reference image, the following columns present segmentation results. In the second column of Figure~\ref{fig:qualitative}, Feature 3DGS produces subpar results, as it lacks feature compactification after SAM embedding, resulting in loose segmentations. The third column shows SAGA, which enforces affinity compactness but fails to handle transient occlusions, producing floating and spiky 3D Gaussians, as seen in Rows 1 and 2. The fourth column displays GS-W with projection-based segmentation, which struggles to eliminate background Gaussians behind target objects. In contrast, our method in the fifth column handles in-the-wild scenes more effectively. We encourage readers to check the supplementary material for more experimental results.

\begin{table}[h]
\centering
\setlength{\tabcolsep}{4pt} 
\renewcommand{\arraystretch}{1}
\caption{Ablation study on Seg-Wild using three scenes from the PT dataset~\cite{pt}. ``aff. feat. ch.'' denotes the number of affinity feature channels.}
\begin{tabular}{ccccccc}
\toprule
& \multicolumn{2}{c}{BG} & \multicolumn{2}{c}{TM} & \multicolumn{2}{c}{TF} \\
\cline{2-7}
& IoU$\uparrow$ & Acc$\uparrow$ & IoU$\uparrow$ & Acc$\uparrow$ & IoU$\uparrow$ & Acc$\uparrow$ \\
\hline
(1) w/o $L_{Com}$ & 0.882 & 0.899 & 0.801 & 0.879 & 0.803 & 0.897 \\
(2) w/o SASM & 0.891 & 0.924 & 0.812 & 0.899 & 0.812 & 0.912 \\
(3) w/o sky mask & 0.910 & 0.953 & 0.824 & 0.908 & 0.810 & 0.909 \\
(4) w/o SGC & 0.871 & 0.934 & 0.797 & 0.903 & 0.795 & 0.912 \\
(5) $\tau$ = 0.3 & 0.924 & 0.955 & 0.839 & 0.916 & 0.816 & 0.919 \\
(6) $\tau$ = 0.7 & 0.919 & 0.951 & 0.835 & 0.913 & 0.813 & 0.915 \\
(7) aff. feat. ch. 32 & 0.912 & 0.937 & 0.833 & 0.911 & 0.817 & 0.913 \\
(8) aff. feat. ch. 64 & 0.930 & 0.968 & 0.846 & 0.920 & 0.827 & 0.924 \\
(9) aff. feat. ch. 128 & 0.931 & 0.968 & 0.847 & 0.920 & 0.826 & 0.930 \\
\midrule
(10) Complete model & 0.930 & 0.968 & 0.846 & 0.920 & 0.827 & 0.924 \\
\bottomrule
\end{tabular}
\label{tab:ablation}
\end{table}

\section{Ablation Studies and Analysis}
We verified the effectiveness of the design of our method on three scenes of the Photo Tourism(PT) dataset~\cite{pt}. Table~\ref{tab:ablation} shows the quantitative results.

\subsection{3D Feature Field Reconstruction}
Row 1 of Table~\ref{tab:ablation} shows the segmentation results without the loss term $L_{Com}$ in 3D feature field reconstruction. $L_{Com}$ uses SAM-generated masks to guide the classification and aggregation of affinity features in the loose 3D feature field. This enhances correlations among 3D Gaussians in the scene. The results show that removing $L_{Com}$ significantly reduces segmentation performance, confirming the effectiveness of our design.

Rows 2–3 of Table~\ref{tab:ablation} illustrate the impact of the scale-adaptive segmentation module (SASM) on the performance of Seg-Wild. In Row 2, removing SASM results in a noticeable performance drop across all three scenes. This decline is primarily due to the fixed segmentation scale used by SAM, which yields inconsistent segmentation across different images. This inconsistency hinders feature alignment across training iterations. In Row 3, removing the sky mask in SASM slightly reduces performance. The sky mask mitigates the influence of the vast background depth typically associated with sky regions, helping to maintain a more accurate distribution of prompt points for foreground segmentation.

\begin{figure}[H]
\centering
\includegraphics[width=0.7\columnwidth]{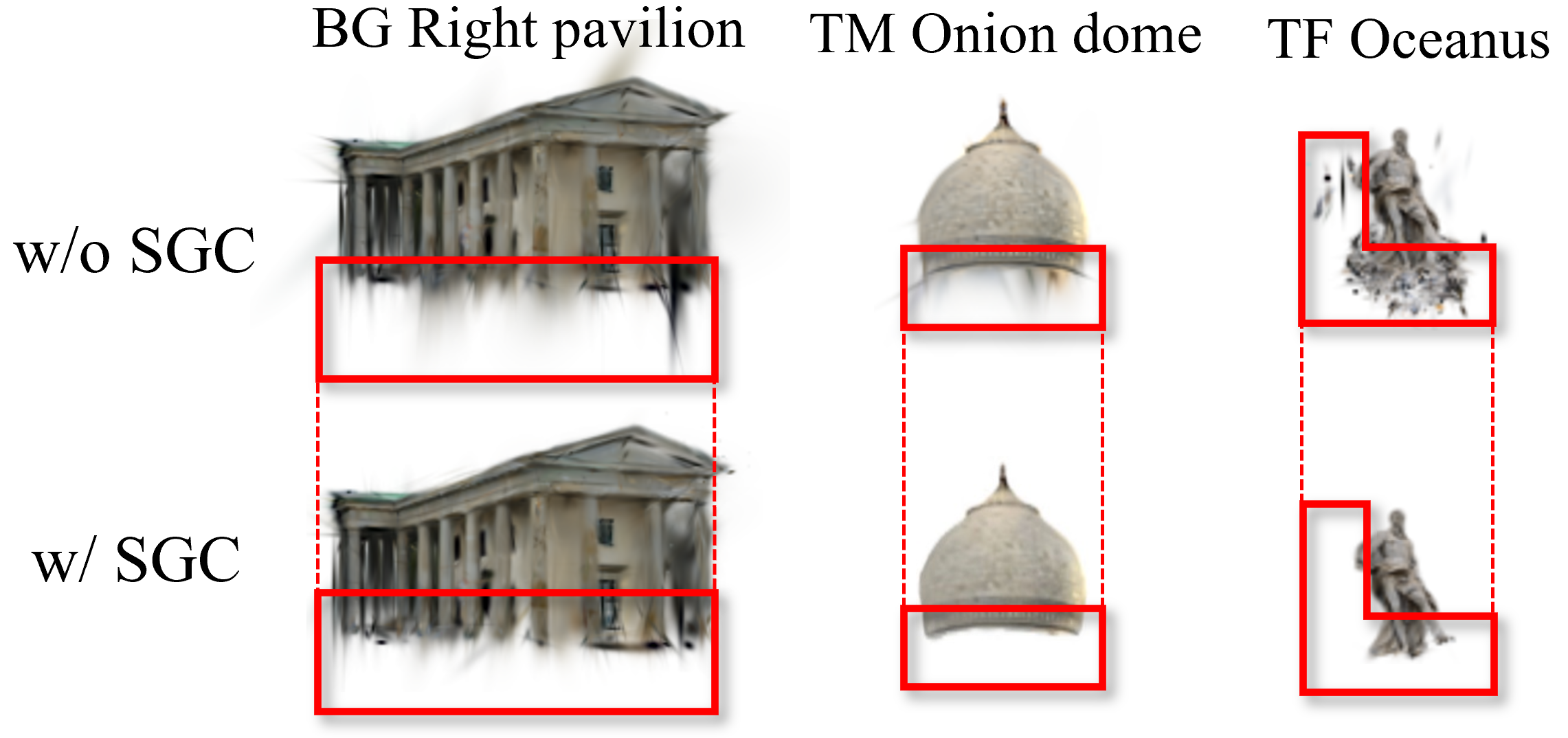}
\caption{Visualization of segmentation results before and after applying the Spiky 3D Gaussian Cutter (SGC). The SGC post-processing effectively removes erroneous and protruding 3D Gaussians, leading to smoother and more coherent segmentation boundaries.}
\label{fig:ablationSGC}
\end{figure}

\subsection{Post-processing}
The primary post-processing technique employed by Seg-Wild is the Spiky 3D Gaussian Cutter (SGC). Row 4 of Table~\ref{tab:ablation} demonstrates that the segmentation results exhibit an average decrease of 4.6\% in IoU and 1.9\% in Accuracy. This reduction highlights the significant impact of the SGC pruning mechanism, which plays a crucial role in enhancing the overall segmentation performance. The visualization results, presented in Figure~\ref{fig:ablationSGC}, provide a more precise illustration of the functionality and effectiveness of SGC.

\subsection{Affinity Feature Channels}
Rows 5-7 of Table~\ref{tab:ablation} present segmentation results using varying numbers of affinity feature channels. These channels represent the feature dimensions of the affinity features stored in each 3D Gaussian within the 3D feature field. More channels allow richer feature information to be preserved. The results show that increasing the number of channels improves segmentation performance. The ablation study results indicate that increasing the number of affinity feature channels improves segmentation performance. However, it also leads to a significant increase in computational complexity. Thus, we select 64 channels as a trade-off between performance and computational efficiency.

\section{Conclusion}
This paper proposes a 3D segmentation method for scenes reconstructed from unconstrained photo collections, enabling efficient and accurate segmentation based on 3DGS. We design several key modules in the pipeline to enhance segmentation performance. We first generate a 3D feature field by lifting 2D feature embeddings into 3D affinity features. We then present a scale-adaptive segmentation module (SASM) to improve the segmentation consistency of SAM in unconstrained photo collections. Finally, we introduce a spiky 3D Gaussian cutter (SGC) as a post-processing step. By integrating 2D SAM masks, SGC removes spiky 3D Gaussians and smooths the segmentation results. Experiments show that our method achieves high-quality segmentation for in-the-wild scenes reconstructed by 3DGS. Meanwhile, our method preserves lighting conditions and color tones consistent with the reference images. In addition, we discuss the failure cases in the supplementary material.

\bibliographystyle{ACM-Reference-Format}
\bibliography{main}






\clearpage 
\appendix  

In the supplemental material, we will provide a more in-depth demonstration of our method. In Section A, we will supplement the implementation details of the technique. In Section B, we will explain and present the implementation details of our experiments in detail. In Section C, we will showcase more experimental results. In Section D, we will discuss the failure cases and future work.

\section{More Details of the Method}
\subsection{3DGS}
We provide detailed descriptions of the computational formulas related to 3D Gaussian Splatting (3DGS)~\cite{3dgs}. 3DGS represents a static scene using a set of 3D anisotropic Gaussians and achieves real-time image rendering through a differentiable tile-based rasterizer. These 3D Gaussians $G(x)$ are represented by a three-dimensional covariance matrix $\Sigma\in\mathbb{R}^{3x3}$ and a three-dimensional central point position $p\in\mathbb{R}^3$:
\begin{equation}
G\left(x\right)=e^{-\frac{1}{2}{(x-p)}^T\Sigma^{-1}(x-p)}\ .\ \
\label{eq1}
\end{equation}

The three-dimensional covariance matrix $\Sigma$ can be calculated through a rotation matrix $R\in\mathbb{R}^{3\times3}$ and a scaling matrix $S\in\mathbb{R}^{3\times3}$. The calculation formula is as follows:
\begin{equation}
\Sigma=RSS^TR^T,
\label{eq2}
\end{equation}

\begin{equation}
\Sigma' = J W \Sigma W^T J^T,
\label{eq3}
\end{equation}

\begin{equation}
p' =\ JWp,\ 
\label{eq4}
\end{equation}
$J$ represents the Jacobian of the projection affine approximation, $W$ is the viewing transformation, and $\Sigma'$ is the covariance matrix in the camera coordinate system. $p'\in\mathbb{R}^2$ represents the corresponding coordinate of $p$ projected onto the 2D plane.

\subsection{2D Features to 3D Features}
In Section 4.1, the calculation formula for $S({px}_i,{px}_j)$ is presented as follows:
\begin{equation}
S({px}_i,{px}_j)=\frac{\left|M_{{px}_i}\cap M_{{px}_j}\right|}{\left|M_{{px}_i}\cup M_{{px}_j}\right|+\epsilon},
\label{eq11}
\end{equation}
where $\epsilon$ is set to ${10}^{-5}$ to prevent division by zero.

In our 3D Gaussian data structure $G_i$, we draw inspiration from GS-W~\cite{gsw} to introduce two features, ${sf}_i$ and ${df}_i$. ${sf}_i$ encodes intrinsic traits from material and surface properties, which remain stable across environments. ${df}_i$ captures dynamic traits influenced by lighting, weather, and shadows. Traditional methods often fail to (1) capture local dynamics via global features and (2) separate ${sf}_i$ and ${df}_i$. We separate them to better model appearance changes specific to in-the-wild scenes and enables editing by modifying ${df}_i$ as shown in Figure~\ref{fig:appearance}.

\subsection{Feature compression}
In Section 4.1, to reduce computational cost, we apply Principal Component Analysis (PCA) to compress the 256-dimensional ${fe}_{SAM}$ features into 64 dimensions. These compressed features are then embedded into 3D Gaussians. To ensure accuracy, a CNN encoder is jointly trained with the 3D Gaussian embedding features to decompress the 64-dimensional features back to 256 dimensions, aligning the affinity features with the original embedding space. 

The CNN decoder contains a single 1$\times$1 convolutional layer that performs a linear projection of input feature maps from 64-dimensional to 256-dimensional channels, preserving the spatial dimensions.

\subsection{Scale-Adaptive Segmentation Module (SASM)}
We provide the calculation formula for the segmentation scale ${SegS}_I$ as follows:
\begin{equation}
{SegS}_I=\left\lfloor s_{min}+norm(d_I)({s_{max}-s}_{min})+0.5\right\rfloor,
\label{eq16}
\end{equation}
where $norm(\cdot)$ represents normalization, $\left\lfloor\cdot\right\rfloor$ represents rounding down, $s_{max}$ and $s_{min}$ are set to 4 and 8, respectively, representing the upper and lower limits of the segmentation scale.

\subsection{Spiky 3D Gaussian Cutter (SGC)}
$\Sigma_{uv}^i\in\mathbb{R}^{2\times2}$ represents the covariance matrix on the two-dimensional plane, which is computed using the following formula:
\begin{equation}
{\Sigma_{uv}^i=J\Sigma}^iJ^T.
\label{eq29}
\end{equation}

\begin{table}[ht]
\centering
\caption{The photos are selected from the PT dataset in the benchmark. The abbreviations ``BG'', ``TM'', and ``TF'' respectively refer to the ``Brandenburg Gate'', the ``Taj Mahal'', and the ``Trevi Fountain''.}
\resizebox{\linewidth}{!}{
\renewcommand{\arraystretch}{0.9}
\begin{tabular}{l l l}
\toprule
\textbf{Segmentation Name} & \textbf{Reference Image} & \textbf{Ground Truth} \\
\midrule
BG Lintel & 00315862\_6836283050.jpg & 31464383\_1528913803.jpg \\
&  & 24431464\_4389038015.jpg \\
&  & 47042487\_3641229825.jpg \\
BG Right pavilion & 96519477\_11703946585.jpg & 80460739\_5924637395.jpg \\
&  & 99002117\_2965987139.jpg \\
&  & 22689144\_8063124816.jpg \\
BG Left pavilion & 29877554\_3225783747.jpg & 24491197\_4983689000.jpg \\
&  & 47405748\_345277775.jpg \\
&  & 01069771\_8567470929.jpg \\
TF Oceanus & 00657038\_9404825385.jpg & 00234320\_6263685886.jpg \\
&  & 01081603\_5459971228.jpg \\
&  & 02305018\_4000998904.jpg \\
TF Triton & 00657038\_9404825385.jpg & 12017941\_12586787884.jpg \\
&  & 21324671\_251787492.jpg \\
&  & 00446387\_2504417907.jpg \\
TF Left hippocampus & 15023987\_46718086.jpg & 12017941\_12586787884.jpg \\
&  & 09080614\_2963736516.jpg \\
&  & 36294258\_3929962941.jpg \\
TM Onion dome & 00077913\_6801374468.jpg & 08564421\_4437928438.jpg \\
&  & 11201698\_2876822374.jpg \\
&  & 18431346\_3891005633.jpg \\
TM Left minaret & 35591645\_2236971918.jpg & 23859155\_1447291459.jpg \\
&  & 01106378\_7955217644.jpg \\
&  & 27054285\_8699868579.jpg \\
TM Right chhatris & 00570394\_12923626565.jpg & 03279294\_1600450773.jpg \\
&  & 25325322\_2724107819.jpg \\
&  & 31426149\_2272885965.jpg \\
\bottomrule
\end{tabular}
}
\label{tab:benchmark}
\end{table}

\section{Experimental Implementation}
\subsection{Details of Benchmark}
We followed mainstream segmentation benchmarks~\cite{nvos, spinnerf}, selected suitable reference images and ground truth annotations from the PhotoTourism (PT) dataset~\cite{pt}, and built a benchmark for evaluating segmentation in the wild. For benchmark generation, we employed the open-source ISAT framework in conjunction with the Segment Anything Model (SAM). Specifically, we used SAM’s built-in prompt-based segmentation to generate initial 2D masks. These masks were manually refined to improve ground truth quality and ensure reliable evaluation. The benchmark contains three target objects in three scenes, and the specific contents are shown in Table~\ref{tab:benchmark}.

We evaluate performance using Intersection over Union (IoU) and Accuracy (Acc). These metrics are computed by comparing the rendered masks with the 2D ground truth masks.


\subsection{Details of the Segmentation Scale Parameter Settings}
\begin{figure}[h]
    \centering
    \includegraphics[width=\linewidth]{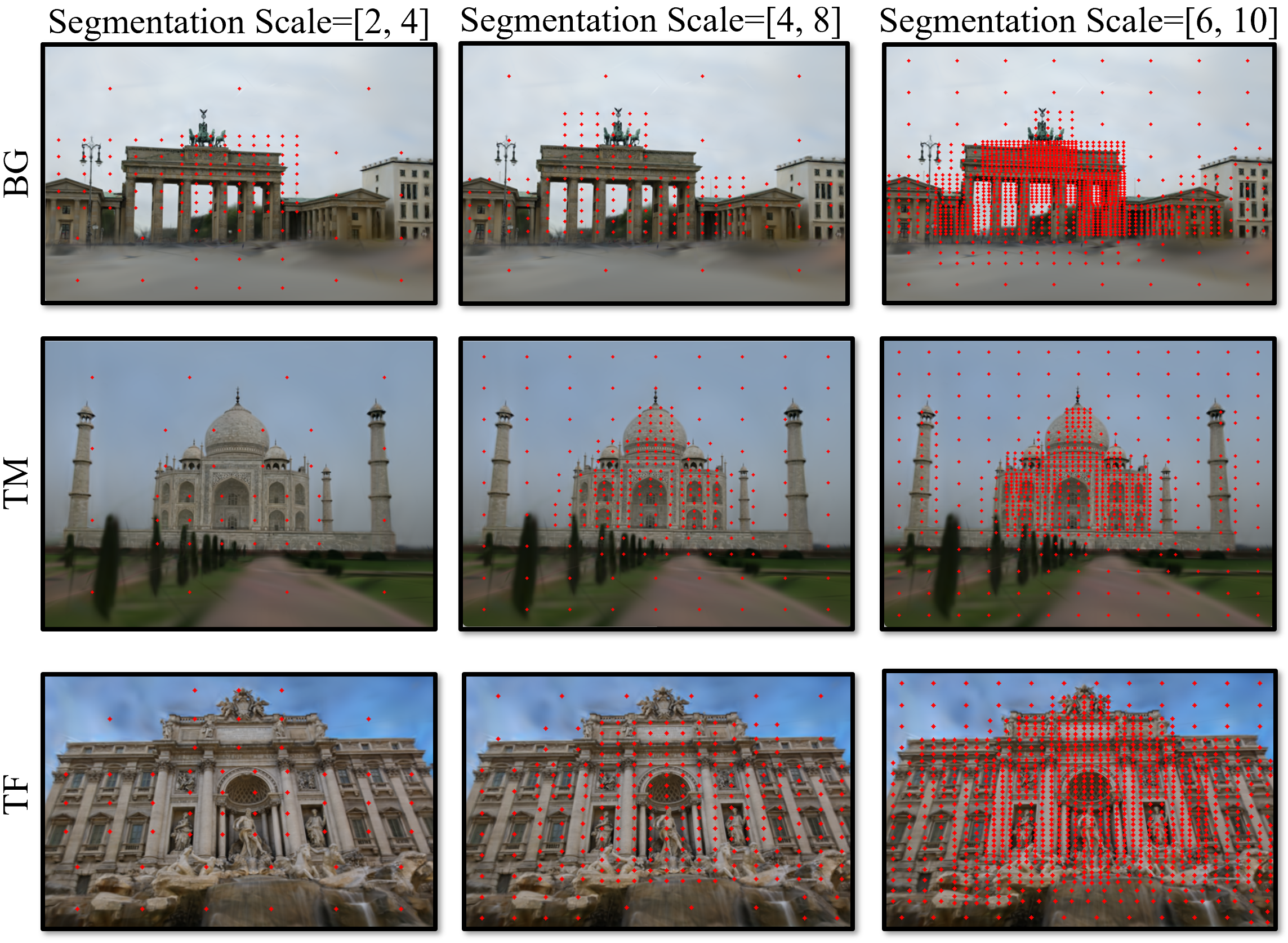}
    \caption{Visualization of the sampling points generated with different Segmentation Scale ${SegS}_I$ values. The abbreviations ``BG'', ``TM'', and ``TF'' refer to the ``Brandenburg Gate'', the ``Taj Mahal'', and the ``Trevi Fountain'', respectively.}
    \label{fig:sample_points}
\end{figure}

In the Scale-Adaptive Segmentation Module (SASM), we set the Segmentation Scale ${SegS}_I$ to [4, 8], the optimal normalization range derived from our experiments. When the number of points is set too large, it generates a significant number of sampling points, leading to a substantial computational load for the input Segment Anything Model (SAM) and impacting efficiency. If the number of points is too small, the generated sampling points will be too sparse, leading to unsatisfactory segmentation results. As shown in Figure~\ref{fig:sample_points}, it is a visualization of the sampling points generated by different Segmentation Scales ${SegS}_I$.

\subsection{Details of the Sky Mask}
\begin{figure}[h]
    \centering
    \includegraphics[width=0.9\linewidth]{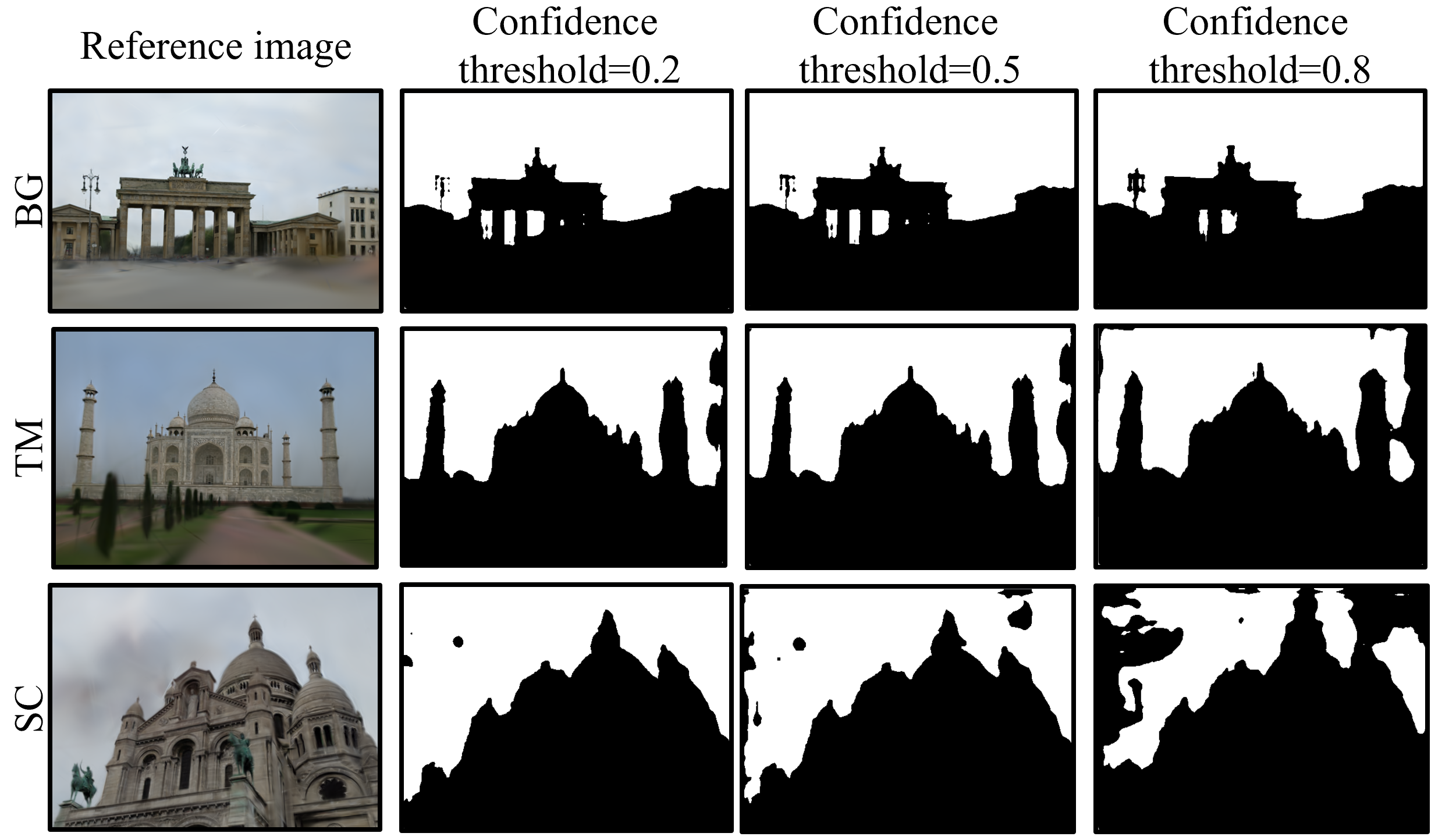}
    \caption{Visualization of sky masks under different confidence thresholds. The abbreviations ``SC'' refer to the  ``Sacre Coeur''.}
    \label{fig:sky_mask}
\end{figure}

We generate sky masks using various pre-trained visual models, such as DeepLabv3+~\cite{deeplabv3+}, LSeg~\cite{lseg}, and SEEM~\cite{seem}. In this paper, we use the version with MobileNet~\cite{mobilenets} as the backbone network and the Cityscapes dataset~\cite{cityscapes} for pre-training, with the "output\_stride" parameter set to 16. In the Cityscapes dataset, the category index for the sky is 10. Therefore, we can obtain the sky mask by having DeepLabv3+ output the segmentation result for the 10$th$ category. As shown in Figure~\ref{fig:sky_mask}, we compared the output results of DeepLabv3+ under different confidence threshold conditions. We aim to achieve a good segmentation result while preserving details, so we set the confidence threshold to 0.5. The segmentation results from pre-trained vision models serve only as a prior reference, and minor segmentation errors do not significantly affect the performance of our method.

\begin{figure}[h]
    \centering
    \includegraphics[width=1\linewidth]{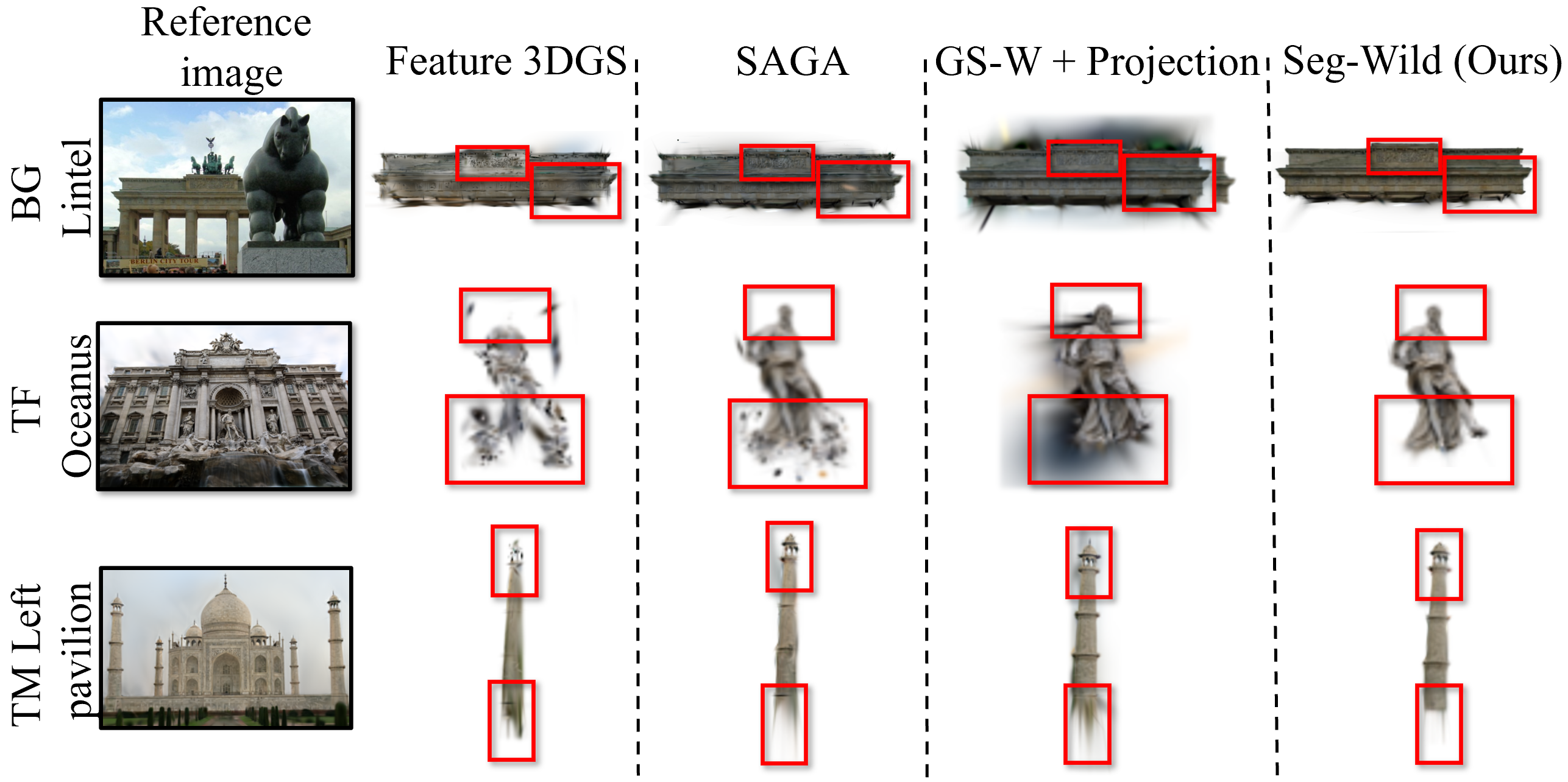}
    \caption{Qualitative comparison of segmentation results across four methods: Feature 3DGS~\cite{feature3d}, SAGA~\cite{saga}, GS-W~\cite{gsw} with projection-based segmentation, and our proposed approach.}
    \label{fig:more_qualitative}
\end{figure}

\section{More Experimental Results}
\subsection{Comparative Experimental Results}
Our qualitative experiments compare three methods: Feature 3DGS\\~\cite{feature3d}, SAGA~\cite{saga}, and GS-W~\cite{gsw} with projection-based segmentation. Figure~\ref{fig:more_qualitative} shows the visualized segmentation results. Feature 3DGS fails to preserve many foreground 3D Gaussians due to the lack of feature compaction. As shown in the second column of Figure~\ref{fig:more_qualitative}, although SAGA performs feature compaction, it struggles to remove spiky 3D Gaussians after segmentation, leading to inaccurate results. By integrating projection-based segmentation into GS-W, we mitigate the effects of transient occlusions on the segmentation. However, this approach still fails to eliminate background impurities behind the segmentation projection. In contrast, Seg-Wild effectively addresses all the above issues and achieves precise segmentation results.


\subsection{Novel View Synthesis Results}
\begin{figure}[h]
    \centering
    \includegraphics[width=0.9\linewidth]{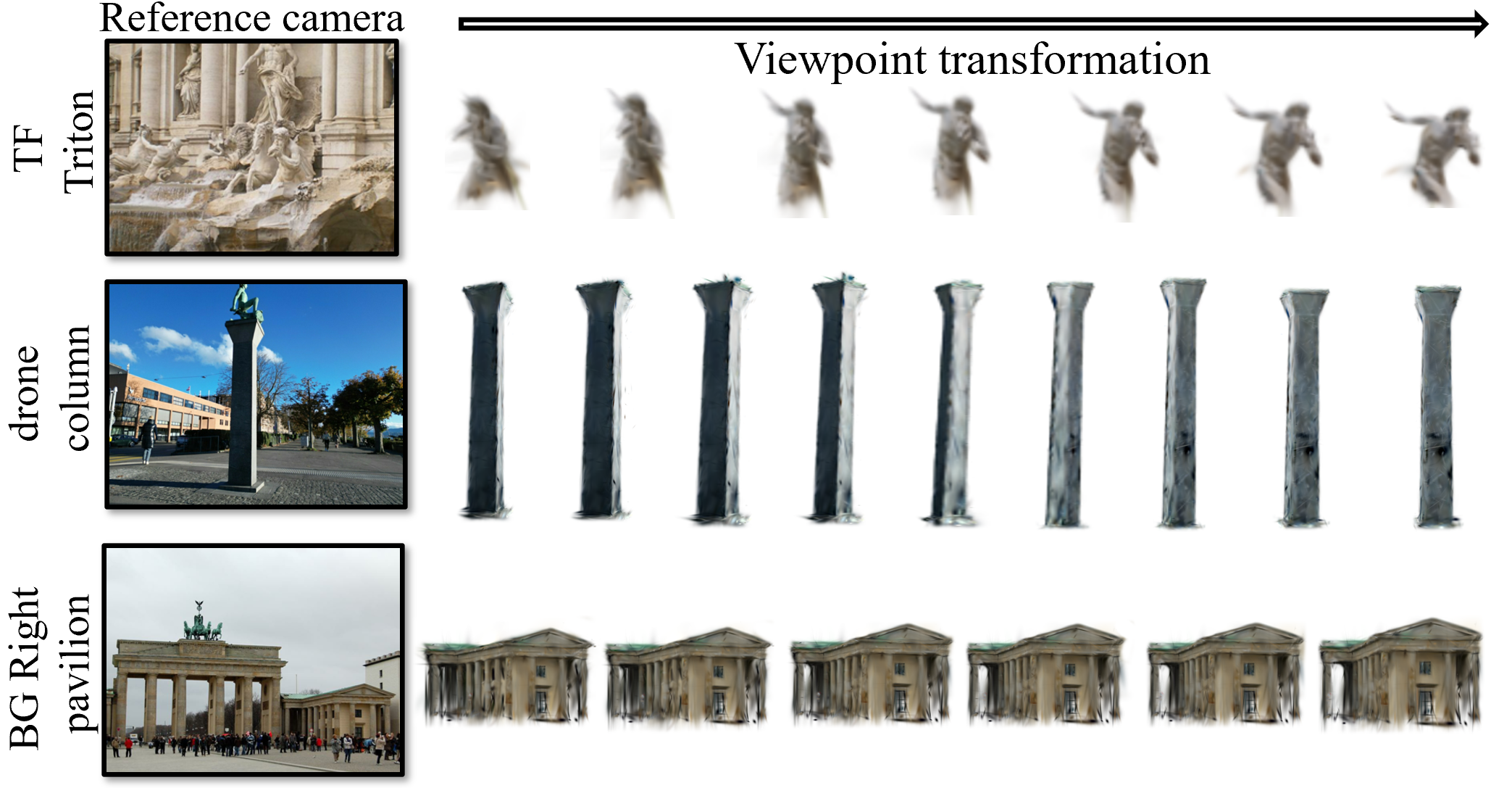}
    \caption{Visualization results of the novel view synthesis for the segmentation.}
    \label{fig:novel_view}
\end{figure}

After completing the segmentation task, we can perform novel view synthesis on the segmentation and render the results using different camera poses. Since our method is based on 3DGS, the speed of novel view synthesis is extremely fast. Figure~\ref{fig:novel_view} presents the detailed visualization results. We show the ``Trevi Fountain'' and the ``Brandenburg Gate'' which are from the PT dataset. Additionally, we also present a ``drone'' from the NeRF-On-the-go dataset~\cite{nerf_on_the_go}.

\subsection{Appearance Tuning Results}
\begin{figure}[h]
    \centering
    \includegraphics[width=0.9\linewidth]{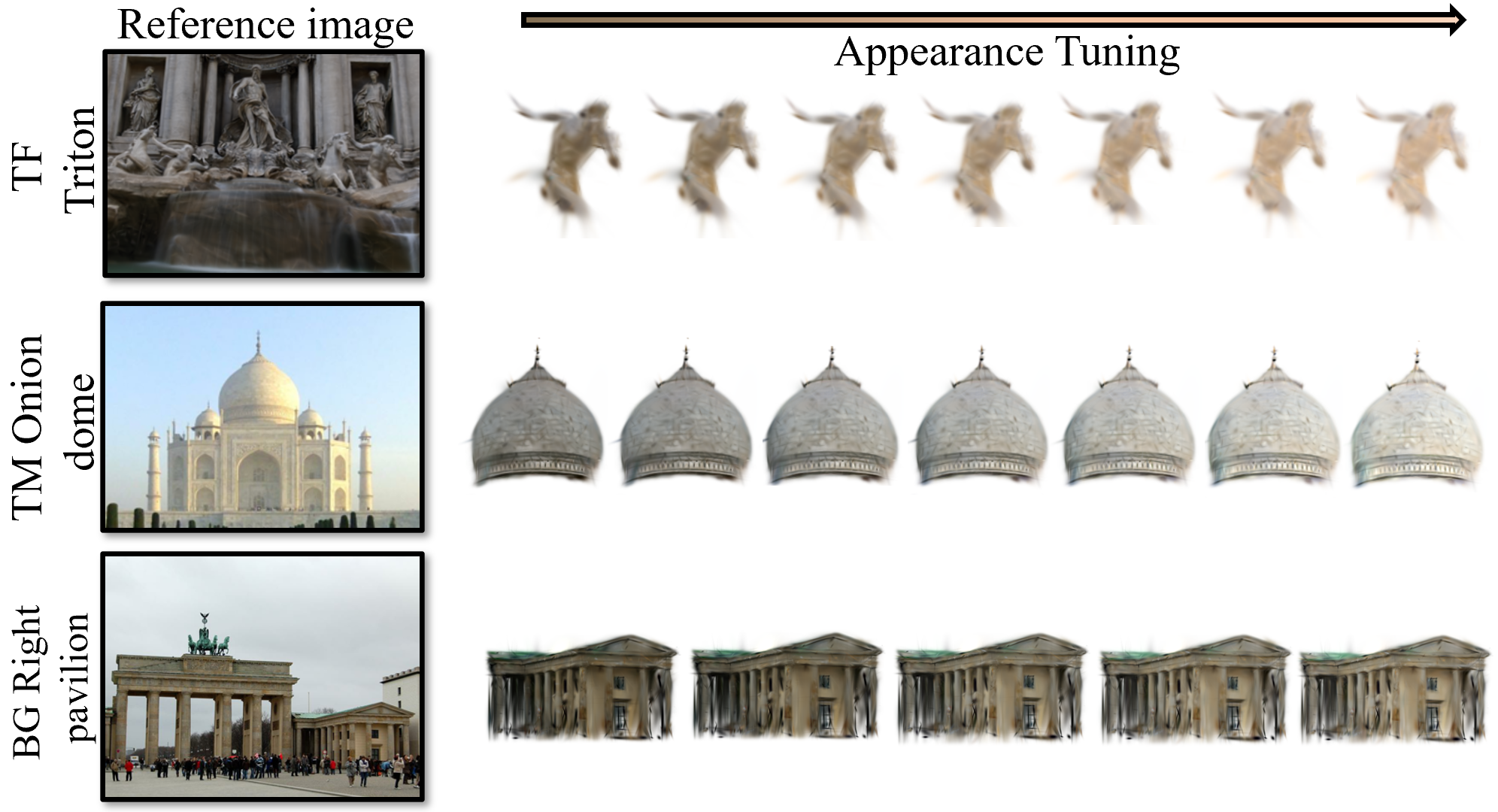}
    \caption{Visualization results of appearance tuning on the segmentation.}
    \label{fig:appearance}
\end{figure}

We perform appearance tuning on the segmented regions by interpolating the appearance weights of two images, allowing us to adjust tone and lighting conditions. Figure~\ref{fig:appearance} shows the corresponding results.

\section{Failure Cases}
\begin{figure}[h]
    \centering
    \includegraphics[width=0.9\linewidth]{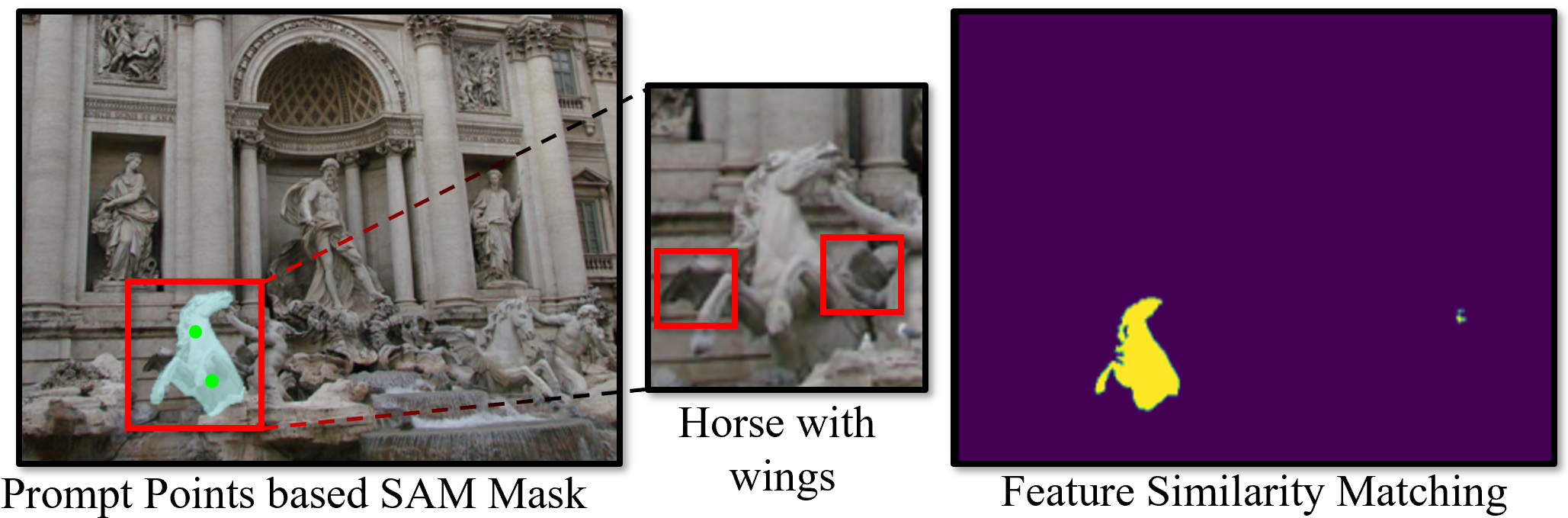}
    \caption{Visualization of Failure Cases. The performance of SAM can slightly affect the segmentation results. A horse with wings is a special case where the SAM mask may be inaccurate, leading to imperfect segmentation.}
    \label{fig:failure_case}
\end{figure}

Although our segmentation method can achieve excellent results in most scenarios, there are still some failure cases. As shown in Figure~\ref{fig:failure_case}, our method encountered problems when attempting to segment the left hippocampus in the ``Trevi Fountain''. The hippocampus is a winged horse in Greek mythology. The Segment Anything Model (SAM)~\cite{sam} is a visual foundation model trained on real-world datasets. Therefore, the SAM Mask segmented through prompt points is highly likely to not include the wings on its back. The inaccuracy of the 2D Mask generated by SAM will slightly affect the performance of our method, and in some exceptional cases, unsatisfactory segmentation results may occur. This issue can be addressed as future researchers continue to improve SAM.

\end{document}